# ADAT: Time-Series-Aware Adaptive Transformer Architecture for Sign Language Translation

Nada Shahin[1,2], Leila Ismail[1,2,3 *]

[1]Intelligent Distributed Computing and Systems (INDUCE) Lab, Department of Computer Science and Software Engineering, College of Information Technology, United Arab Emirates University, Al Ain, Abu Dhabi, United Arab Emirates

[2]National Water and Energy, United Arab Emirates University, Al Ain, Abu Dhabi, United Arab Emirates

[3]Emirates Center for Mobility Research, United Arab Emirates University, Al Ain, Abu Dhabi, United Arab Emirates

The corresponding author's E-mail address: Leila@uaeu.ac.ae

## Abstract

Current sign language machine translation systems rely on recognizing hand movements, facial expressions and body postures, and natural language processing, to convert signs into text. Recent approaches use Transformer architectures to model long-range dependencies via positional encoding. However, they lack accuracy in recognizing fine-grained, short-range temporal dependencies between gestures captured at high framerates. Moreover, their high computational complexity leads to inefficient training. To mitigate these issues, we propose an Adaptive Transformer (ADAT), which incorporates components for enhanced feature extraction and adaptive feature weighting through a gating mechanism to emphasize contextually relevant features while reducing training overhead and maintaining translation accuracy. To evaluate ADAT, we introduce MedASL, the first public medical American Sign Language dataset. In sign-to-gloss-to-text experiments, ADAT outperforms the encoder-decoder transformer, improving BLEU-4 accuracy by 0.1% while reducing training time by 14.33% on PHOENIX14T and 3.24% on MedASL. In sign-to-text experiments, it improves accuracy by 8.7% and reduces training time by 2.8% on PHOENIX14T and achieves 4.7% higher accuracy and 7.17% faster training on MedASL. Compared to encoder-only and decoder-only baselines in sign-to-text, ADAT is at least 6.8% more accurate despite being up to 12.1% slower due to its dual-stream structure.

## 1. Introduction

Sign Language Machine Translation (SLMT) [1] has emerged as a foundational research area in natural language processing (NLP), aiming to reduce communication barriers for the Deaf and hard-of-hearing (DHH) community. With projections indicating that over 700 million people will experience hearing loss by 2050 [2] and the existence of more than 300 distinct sign languages worldwide [3], accessibility remains a significant challenge. The shortage of qualified sign language interpreters [4] further limits real-time communication, particularly in critical situations, such as healthcare and emergency response [5], where effective communication directly impacts individual well-being and safety [6]. Consequently, there is a pressing need to develop scalable, automated, and efficient SLMT systems that facilitate real-time communication while ensuring accuracy, efficiency, and inclusivity.

Recent advancements in machine translation (MT) have improved universal multi-modal representation [7]; however, existing SLMT research presents unique challenges. This is due to the distinct grammatical structures, complex spatiotemporal characteristics, and dynamic visual representations of sign languages. To achieve accurate and efficient translations, SLMT systems must capture short- and long-range dependencies, considering essential sign components such as motion sequences, hand shape and location, facial expressions, and body postures [8]. Nevertheless, as sign language continuously evolves, SLMT models must be frequently retrained to maintain accuracy. Existing models often struggle with this, requiring extensive computational resources and lengthy training time [9]. Furthermore, the scarcity of annotated sign language data remains a fundamental barrier [5], necessitating approaches that can train efficiently while preserving accuracy.

Several SLMT works adopted transformer architectures [5-7], given their state-of-the-art performance in language translation [5]. Transformers effectively model long-range dependencies using positional embeddings and attention mechanisms [13]. However, their quadratic computational complexity poses challenges during training, as they do not efficiently capture fine-grained short-range temporal dependencies [10, 11] – an essential aspect for processing high-frame-rate data, typically recorded at 30-60 frames per second. These inefficiencies limit the scalability of SLMT systems, particularly in frame-by-frame analysis, making it challenging to develop real-time applications and rapidly adapt to new data.

In this study, we fill this void by proposing an Adaptive Transformer (ADAT), a novel architecture designed to enhance SLMT by dynamically capturing the fine-grained short- and long-range spatiotemporal features while improving training speed. This is by integrating multiple key components: convolutional layers to extract localized sign features, LogSparse Self-Attention (LSSA) [16] to reduce computational overhead by attending to log-spaced past sign frames, and an adaptive gating mechanism [17] to selectively retain critical temporal dependencies. By combining these elements, ADAT effectively captures short- and long-range temporal dependencies and improves training efficiency, allowing faster adaptation to new data and evolving sign languages. Compared to the transformer proposed by Vaswani et al. [13], ADAT optimizes attention computations and reduces unnecessary overhead, facilitating more effective sign language translation.

We divide the SLMT system into two core processes: 1) Sign Language Recognition (SLR), known as sign-to-gloss (S2G), which converts video sequences of signs into glosses that are symbolic representations that preserve sign language linguistics, and 2) Sign Language Translation (SLT), which generates spoken text from the recognized signs. SLT can be achieved through direct sign-to-text (S2T) translation or sign-to-gloss-to-text (S2G2T) translation [1].

We evaluate ADAT for S2T and S2G2T in comparison with three transformer baselines: encoder-decoder, encoder-only, and decoder-only. Our evaluation utilizes two benchmark datasets: the RWTH-PHOENIX-Weather-2014 (PHOENIX14T) [1] and MedASL, our medical American Sign Language (ASL) dataset, addressing the critical need for accurate translation in healthcare where communication barriers can have significant consequences on the patient safety and well-being [6]. Our results highlight the potential of ADAT in advancing SLMT systems and closing the communication gap between the Deaf community and broader society, particularly in crucial sectors such as healthcare [18].

The main contributions of this paper are as follows:

- We propose an Adaptive Transformer (ADAT), a novel transformer-based model that dynamically captures short- and long-range temporal dependencies while optimizing computational efficiency.
- We perform a comparative evaluation of ADAT with encoder-decoder, encoder-only, and decoder-only transformer baselines. The experimental results show that ADAT outperforms the baselines in terms of accuracy and training time for S2G2T and S2T.
- We evaluate ADAT using two datasets of different characteristics and sign languages: the largest public German weather-related PHOENIX14T dataset and MedASL, a novel continuous ASL medical-related dataset that we introduce.

The rest of the paper is organized as follows: Section 2 overviews the related work. The proposed adaptive transformer is described in Section 3. Section 4 discusses the proposed MedASL dataset. Section 5 presents the experimental setup. Numerical experiments and comparative performance results are provided in Section 6. Section 7 concludes the paper with future research directions.

## 2. Related Works

We categorize the related works into two categories: 1) S2G2T and 2) S2T, summarized in Table 1.

Several works explored S2G2T [1,10,19–25]. [1] introduced the first end-to-end S2G2T model based on neural machine translation. Their model learned spatio-temporal sign representations while simultaneously mapping glosses to spoken language. While this work was a foundational step forward, it was limited by the reliance on pre-trained AlexNet, domain-specific evaluation on PHOENIX14T, and lack of time complexity consideration. [10] applied the encoder-decoder transformer with pre-trained EfficientNets and connectionist temporal classification (CTC) loss. This approach enabled gloss recognition without explicit frame-to-gloss alignment, as CTC loss introduces blank tokens to allow flexibility in timing [26], thereby improving performance. However, the model evaluation was limited to the PHOENIX14T dataset, restricting its generalizability. [19] introduced the Sign Back-Translation model to overcome sign language data scarcity by generating sign sequences from monolingual text. While this improved model training, it relied on pre-trained CNN-based feature extraction and. [20] employed a graph-based model that treated body parts as spatio-temporal nodes, using ResNet-152 for feature extraction. However, unlike transformer-based SLMT, this approach separated S2G and G2T, requiring a multi-step process that prevented direct translation optimization, leading to suboptimal performance. [21] proposed PiSLTRc, a transformer-based system that integrates pre-trained EfficientNets with content- and position-aware temporal convolutions. Depending on transfer learning between stages, their approach required separate training for S2G and Gloss2Text (G2T). This prevented backpropagation from text outputs to sign video inputs, limiting translation accuracy. [22] applied contrastive learning to enhance the visual and semantic robustness. This approach introduced higher computational overhead, making S2G2T training significantly more resource-intensive. [23] introduced SLTUNET, a multi-task learning framework integrating S2G, G2T, and S2T in a single model to share representations across tasks. While they employed transfer learning, task interference would lead to longer fine-tuning time as errors in gloss recognition impact text translation. [25] applied pre-trained S3D for feature extraction, followed by a lightweight head network for temporal feature processing, along with CTC loss for S2G. They then fed the predicted glosses into a G2T transformer-based model, where the performance of S2G and G2T was optimized individually. The results demonstrated that glosses lose spatio-temporal visual information as they are simplified written sign language that only captures word-level meaning without the crucial aspects of sign language, such as facial expression and non-manual markers [8]. Therefore, any inaccuracy in S2G can be propagated to G2T, leading to lower performance. [24] proposed SimulSLT, an end-to-end simultaneous SLT model using a masked transformer encoder and a wait-k strategy. While this approach reduced latency, it depended on pre-trained models, requiring significant computational resources.

In summary, several works on S2G2T adopted transformer-based models. However, these works depend on pre-trained models and complex computations. Future research should consider diverse datasets, independence from transfer learning, and improve computational efficiency.

Furthermore, several works investigated S2T translation [1,10,19,21,23–25,27–32]. While this approach eliminates gloss annotations, it increases complexity due to the need for long-range dependencies across sign sequences and text outputs. This leads to challenging and impractical real-world SLMT system deployment compared to S2G2T translation [1,10,19,21,23–25]. However, [25] proposed a visual-language mapper to retain spatio-temporal information from sign videos that glosses cannot represent. [27] introduced an SLMT framework, incorporating word existence verification, conditional sentence generation, and cross-modal re-ranking. While this improved translation accuracy, it relied on pre-trained language models. [28] proposed a transformer-based model utilizing linguistic memory storage to improve sentence-level accuracy. However, this approach depended on external linguistic models. [29] developed SignNet II, a dual-learning transformer-based model that optimizes S2T translation. This approach relied on transfer learning and pre-trained pose-estimation models, increasing training dependencies. [30] introduced a Curriculum-based Non-Autoregressive Decoder to reduce inference latency by generating entire sentences in parallel. However, the training time was not optimized, requiring extensive fine-tuning to maintain accuracy. [31] introduced GASLT, a gloss-free model using gloss attention mechanisms. While this approach removed gloss dependency, it required linguistic knowledge transfer from pre-trained natural language models. [32] introduced another gloss-free model utilizing contrastive language-image pre-training and masked self-supervised learning. While semantic S2T alignment was improved, the model relied on pre-trained models, lacked explicit optimization, and required extended fine-tuning cycles.

In summary, recent works on S2T translation rely on transfer learning and introduce high computational costs. This leads to impractical real-world SLMT deployment compared to S2G2T.

To conclude, transformer-based models dominate SLMT due to their contextual learning capabilities. However, recent works rely on transfer learning approaches due to sign language data scarcity. To our knowledge, there are no existing studies that focus on optimizing training efficiency in SLMT. To fill this void, we propose ADAT, a novel adaptive transformer architecture for sign language translation, to improve training efficiency. In addition, ADAT enhances translation accuracy thanks to its time-series awareness, which takes into consideration the fine-grained short-range dependencies and dynamic contextual characteristics of sign language.

**Table 1:** Comparison between sign language machine translation in literature.

| Work | Sign Language(s) | Dataset(s) | Feature Extraction | Algorithm(s) | Architecture Modification | Transfer Learning | Training time | Translation Tasks: S2G2T | Translation Tasks: S2T |
|---|---|---|---|---|---|---|---|---|---|
| [1] | DGS | PHOENIX14T | AlexNet | LSTM & GRU* | Attention layer | ✔ | ✗ | ✔ | ✔ |
| [10] | DGS | PHOENIX14T | EfficientNets | Transformer | None | ✔ | ✗ | ✔ | ✔ |
| [19] | DGS & CSL | CSL-Daily & PHOENIX14T* | Pre-trained CNN | Transformer | • Temporal Inception Network<br>• Batch normalization layer | ✔ | ✗ | ✔ | ✔ |
| [20] | DGS & CSL | CSL-Daily & PHOENIX14T* | ResNet-152 | Graph Neural Network | • Convolution<br>• Self-attention<br>• Pooling | ✔ | ✗ | ✔ | ✗ |
| [21] | CSL & DGS | CSL, PHOENIX14, & PHOENIX14T* | EfficientNets | Transformer | • Content- & position-aware temporal convolution for feature selection<br>• Positional encoding is replaced by Disentangled Relative Position Encoding<br>• Content-aware self-attention | ✔ | ✗ | ✔ | ✔ |
| [22] | CSL & DGS | CSL-Daily & PHOENIX14T* | ResNet18 | Transformer | • Contrastive learning<br>• Temporal Module<br>• No decoder in the recognition module | ✔ | ✗ | ✔ | ✗ |
| [23] | CSL & DGS | CSL-Daily & PHOENIX14T* | SMKD model | Transformer | • Separate encoders for video and text before merging into a shared encoder<br>• Multiple SLT tasks are trained simultaneously | ✔ | ✗ | ✔ | ✔ |
| [24] | DGS | PHOENIX14T | AlexNet | Transformer | • Masked encoder<br>• Wait-k and auxiliary decoders<br>• Boundary predictor | ✔ | ✗ | ✔ | ✔ |
| [25] | DGS & CSL | CSL-Daily & PHOENIX14T* | S3D | Transformer | • Visual-Language Mapper | ✔ | ✗ | ✔ | ✔ |
| [27] | CSL & DGS | CSL & PHOENIX14T* | CNN & transformer encoder | Transformer | • No positional encoding<br>• Decoding using BERT<br>• Cross-modal re-ranking for final prediction | ✔ | ✗ | ✗ | ✔ |
| [28] | DGS | PHOENIX14T | CNN | Transformer | • Gated Interactive Attention in the encoder<br>• Multi-stream memory structure | ✔ | ✗ | ✗ | ✔ |
| [29] | ASL & DGS | ASLing & PHOENIX14T* | ResNet50 | Transformer | None | ✔ | ✗ | ✗ | ✔ |
| [30] | CSL & DGS | CSL-Daily & PHOENIX14T* | SMKD model | Transformer | • Non-Autoregressive Decoder<br>• Decoding is bidirectional<br>• Curriculum and Mutual Learning | ✔ | ✔ | ✗ | ✔ |
| [31] | CSL & DGS | CSL-Daily, PHOENIX14T*, SP-10 | I3D | Transformer | Gloss Attention before the Transformer | ✔ | ✗ | ✗ | ✔ |
| [32] | CSL & DGS | CSL-Daily & PHOENIX14T* | • ResNet18<br>• mBART | Transformer | None | ✔ | ✗ | ✗ | ✔ |
| **This Work** | **DGS & ASL** | **PHOENIX14T & MedASL** | **CNN** | **Transformer** | **• Remove the positional encoding and rely on CNN for feature extraction<br>• Splitting the sign video input<br>• CNN and global average in the encoder<br>• Use LogSparse Self-Attention<br>• Adaptive gating in the encoder** | ✗ | ✔ | ✔ | ✔ |

ASL: American Sign Language; CSL: Chinese Sign Language; DGS: German Sign Language; GRU: Gated Recurrent Unit; LSTM: Long Short-Term Memory; *: Best performance; S2G2T: Sign-to-gloss-to-text; S2T: Sign-to-text.

## 3. Proposed ADAT: An Adaptive Time-Series Transformer Architecture

In this section, we consider the SLMT framework that takes sign videos with $F$ frames $x = \{x_f\}_{f=1}^{F}$ as input and translates it into a series of spoken language text with $T$ sequence length as an output $y = \{y_t\}_{t=1}^{T}$. This is while considering gloss annotations with $G$ glosses as an intermediary between $x$ and $y$, $z = \{z_g\}_{g=1}^{G}$. To tackle this, we introduce an Adaptive Transformer (ADAT) to map between sign videos and spoken language text, taking into account the gloss annotations for an efficient SLMT system. Fig. 1 depicts the ADAT architecture, where a) presents an overview of the architecture and b) illustrates the encoder layer in detail. We present Algorithm 1 in Supplementary 1 to elaborate on ADAT's flow.

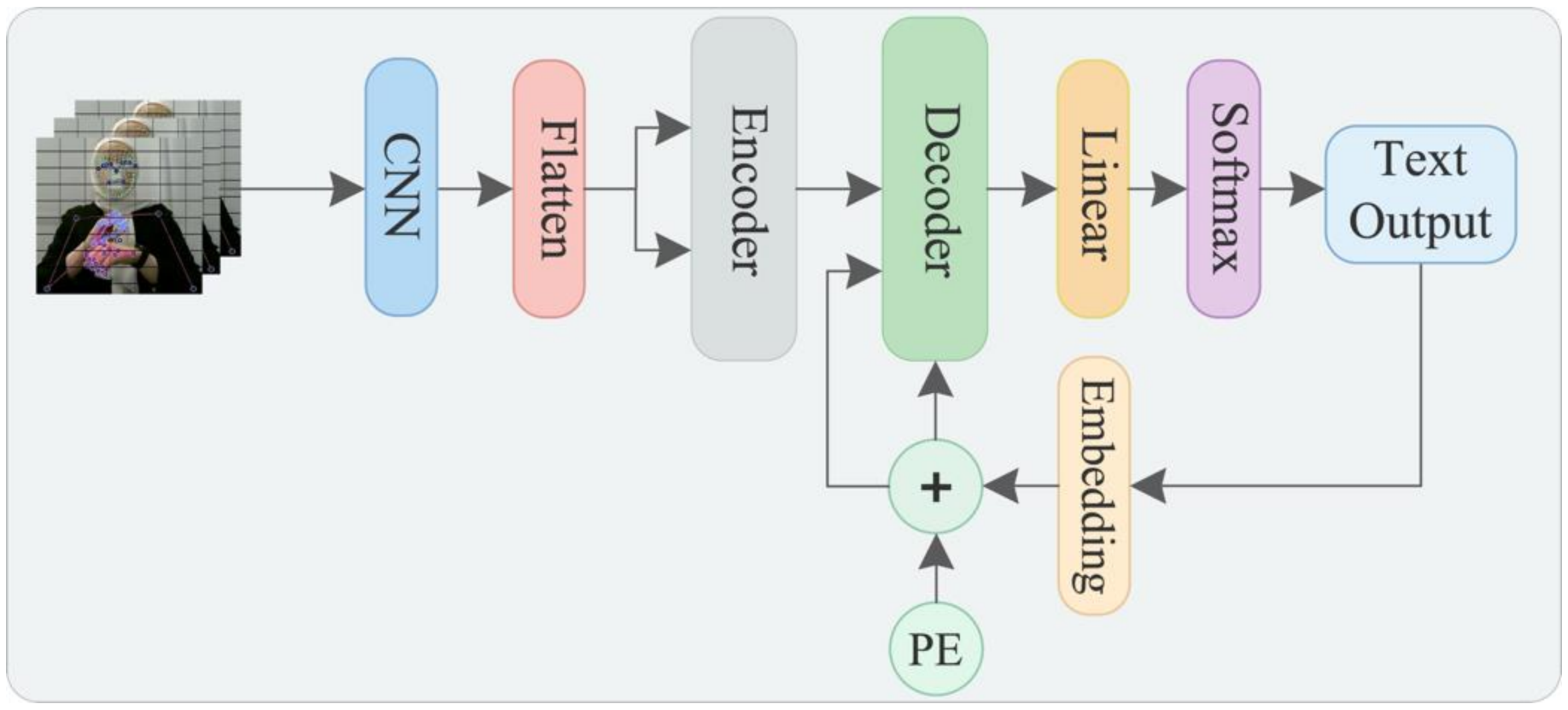

(a)

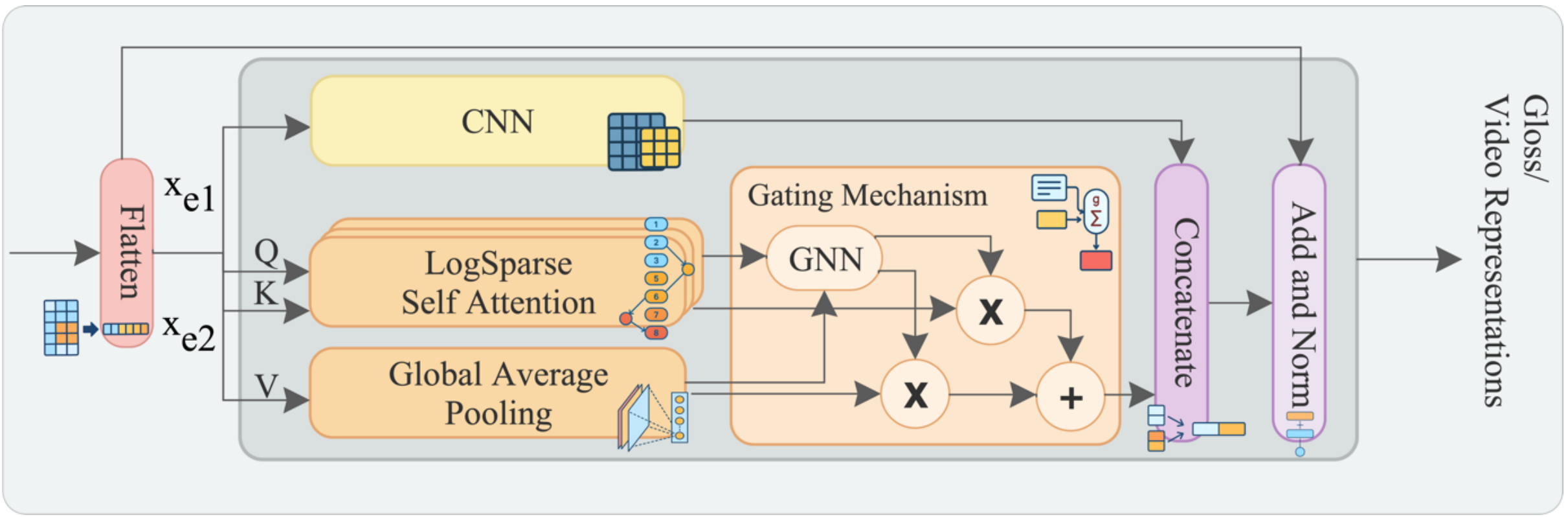

(b)

**Figure 1:** (a) Overall ADAT architecture. (b) ADAT encoder layer, the output is gloss in sign-to-gloss-to-text or video representations in sign-to-text translation. (PE: Positional Encoding, CNN: Convolution Neural Network, GNN: Gated Neural Network. .)

The proposed ADAT architecture is a novel variant of the transformer encoder tailored for SLMT. It integrates convolution layers, LogSparse Self-Attention (LSSA) [16], and an adaptive Gating Mechanism [17] to handle the temporal and dynamic aspects of sign language effectively. This architecture has the following workflow:

### *3.1. Input Feature Extraction and Dimension*

We extract the input features utilizing 2D CNN with 16 filters of size 3×3 and a ReLU activation function, followed by 2×2 max pooling. CNNs are well-suited for capturing local spatial patterns in a heterogeneous environment by transforming the frames into higher-level feature representations, focusing on semantically relevant patterns [33]. Max pooling further refines the feature maps by reducing spatial redundancy, thus optimizing computational efficiency [34].

The input to the encoder consists of sign language videos $x_e$ with dimension 52×65. We choose this resolution to reduce the computational cost while preserving the visual details, ensuring compatibility across datasets of different resolutions. We structure the extracted features as $x_e \in \mathbb{R}^{m\times d}$, where $m$ is the number of frames and $d$ is the feature dimension per frame. $d = C \times H \times W$, where $C = 3$ (RGB channels) and $H \times W$ are 52×65. Then, we flatten $x_e$ for compatibility with subsequent layers.

*3.2. Encoder*

We split the input into two equal parts, each capturing different parts of the sign video, thus reducing computational load. The first part, $x_{e1}$, undergoes a convolutional operation ($Conv$) to extract localized features such as hand shapes and facial expressions. The second part, $x_{e2}$, is processed in two parallel branches: the first is processed using a stacked $LSSA$ to capture long-range dependencies, whereas the second is passed through Global Average Pooling ($GAP$) to compute global dependencies. These components are followed by an adaptive gating mechanism ($Gating$) to selectively integrate essential features [17].

- Self-Attention

The canonical self-attention mechanism [13] calculates attention scores between every input pair, leading to a complexity of $O(L^2)$. This quadratic complexity makes it inefficient for long-sequence modeling and computationally expensive. To mitigate these limitations, we adopt Stacked $LSSA$ [35], inspired by the LogSparse Transformer [16]. Unlike canonical self-attention, which processes all input tokens exhaustively, LSSA selects a logarithmically spaced subset of previous patches with an exponentially increasing step size. This structured sparse attention pattern reduces the complexity to $O(L(logL)^2)$ while maintaining long-range dependencies.

In self-attention [12, 14], input patches are represented as query, key, and value matrices (Q/K/V). However, in stacked $LSSA$, we focus on the Q/K matrices, reducing attention computations while maintaining efficiency, as shown in equation (1).

$$LSSA_{x_{e2}}(Q,K) = Sofmax\left(\frac{Q_{I_p^j} K^T_{I_p^j}}{\sqrt{\frac{d}{2}}}\right) \qquad (1)$$

where $d$ is the dimension and $I_p^j$ is the patch indices to which the current patch $p$ can attend during the computation from $j$ to $J+1$.

This logarithmic sparsity balances the efficiency and performance by dynamically adjusting the receptive field, enhancing long-range pattern recognition while preserving locality. Theorem 1 in Supplementary 2 proves the complexity analysis for the log-sparse self-attention mechanism.

- Global Average Pooling

We incorporate $GAP$ to enhance positional sensitivity by averaging all spatiotemporal information in $x_{e2}$ which are stored in the V matrix derived from Q/K/V [36]. This results in a single value per channel, simplifying the feature map and preserving only the essential global characteristics. equation (2) presents the $GAP$ operation:

$$GAP_{x_{e2},d} = \frac{1}{H\times W}\sum_{i=1}^{H}\sum_{j=1}^{W} V_{i,j} \qquad (2)$$

where $H$and $W$ are the height and width of the feature map and $V_{i,j}$ is the value at position $(i,j)$ of matrix V.

- Adaptive Gating Mechanism

We combine the outputs of $LSSA$ and $GAP$ using an adaptive gating mechanism to balance short- and long-range dependencies. This mechanism contains a gated neural network (GNN) that generates a gate value ($g$) through a SoftMax activation function, forming a probability distribution [37]. Therefore, it dynamically evaluates attention weights and selectively enhances either short- or long-range dependencies [38], depending on the relative importance of the temporal dependencies [17]. We apply the gating mechanism with $LSSA$ and $GAP$, and the convolution of the entire input to replace the static positional encoding and the canonical self-attention, thereby allowing more efficient context-aware dynamic adaptation to varying temporal dependencies. equations (3) and (4) show the gating mechanism formula.

$$g = Softmax(w \cdot LSSA_{x_{e2}} + b) \qquad (3)$$

$$Gating_{x_{e2}} = g \cdot LSSA_{x_{e2}} + (1 - g) \cdot GAP_{x_{e2}} \qquad (4)$$

where $w$ is a learnable weight and $b$ is the bias.

*3.3. Decoder*

The decoder follows the classical transformer decoder structure, consisting of multi-head attention layers, feed-forward layers, and layer normalization [13]. It takes the gloss representations generated by the encoder and autoregressively generates spoken language text.

## 4. Proposed MedASL Dataset

Table 2 presents an overview of the public datasets used in the literature for SLR (S2G), or SLT (S2T or S2G2T). While several datasets consist of video, gloss, and text [1,39–44], they are designed for general-purpose daily communication, with only a few addressing communications in specific domains, such as weather [1, 20], emergency [40], and public service [41]. However, healthcare, a domain where precise communication is critical, remains underrepresented. The lack of medical sign language datasets limits the applicability of SLMT systems in clinical settings, where seamless and accurate communication could improve individual care. Therefore, it is crucial to introduce annotated datasets for healthcare.

Consequently, we introduce MedASL, an ASL corpus which focuses on medical communication, with gloss and text annotations. It is designed to support researchers and industry professionals in advancing SLMT systems. By incorporating medical terminologies and advanced data acquisition, such as the Intel RealSense camera, MedASL enables the development of accurate and context-aware models that reflect real-world healthcare scenarios. The dataset consists of 500 medical and healthcare-related statements, generated via prompt engineering using ChatGPT [45] and signed by an ASL expert, simulating realistic dialogues between patients and healthcare professionals. We provide the prompt engineering design and data pre-processing in Supplementary 3.

**Table 2:** Summary of public sign language translation datasets.

| Dataset | Year | Sign Language | Domain | Video | Gloss | Text | Linguistic Unit | #Videos | Resolution | Acquisition |
|---|---|---|---|---|---|---|---|---|---|---|
| BOSTON-104 [46] | 2008 | ASL | NR | ✔ | ✔ | ✗ | Sentences | 201 | 195×165 | RGB |
| ASLLVD [47] | 2008 | ASL | General | ✔ | ✔ | ✗ | Words | +3,300 | Varies | RGB |
| ASLG-PG12 [48] | 2012 | ASL | NR | ✗ | ✔ | ✔ | Sentences | NA | NA | NA |
| BSL Corpus [42] | 2013 | BSL | General | ✔ | ✔ | ✔ | Sentences | NR | NR | RGB |
| S-pot [43] | 2014 | Suvi | NR | ✔ | ✔ | ✔ | Sentences | 5,539 | 720×576 | RGB |
| PHOENIX14 [39] | 2015 | DGS | Weather | ✔ | ✔ | ✔ | Sentences | 6,841 | 210×260 | RGB |
| BosphorusSign [49] | 2016 | TİD | General | ✔ | ✔ | ✗ | Words & Sentences | +22,000 | 1920×1080 | Kinect v2 |
| CSL [50] | 2018 | CSL | General | ✔ | ✔ | ✗ | Sentences | 5,000 | 1920×1080 | RGB |
| PHOENIX14T [1] | 2018 | DGS | Weather | ✔ | ✔ | ✔ | Sentences | 8,257 | 210×260 | RGB |
| KETI [40] | 2019 | KSL | Emergency | ✔ | ✔ | ✔ | Sentences | 14,672 | 1920×1080 | HD RGB |
| CoL-SLTD [51] | 2020 | LSC | General | ✔ | ✔ | ✗ | Words & Sentences | 1,020 | 448×448 | RGB |
| ASLing [52] | 2021 | ASL | General | ✔ | ✗ | ✔ | Sentences | 1,284 | 450×600 | RGB |
| How2Sign [53] | 2021 | ASL | General | ✔ | ✔ | ✗ | Words & Sentences | +30,000 | 1280×720 | RGB |
| GSL Dataset [41] | 2021 | GSL | Public Service | ✔ | ✔ | ✔ | Sentences | 10,295 | 840×480 | Intel RealSense |
| ISL-CSLTR [44] | 2021 | ISL | General | ✔ | ✔ | ✔ | Sentences | 700 | NR | RGB |
| **MedASL (ours)** | **2025** | **ASL** | **Medical** | ✔ | ✔ | ✔ | **Sentences** | **500** | **1280×800** | **Intel RealSense** |

ASL: American; BSL: British; CSL: Chinese; DGS: German; GSL: Greek; ISL: Indian; KSL: Korean; LSC: Colombian; Suvi: Finnish; TİD: Turkish.
NA: Not applicable; NR: Not reported.

# 5. Performance Evaluation

## *5.1. Experimental Environment*

- Datasets

We evaluate our proposed ADAT model on PHOENIX14T [1] and MedASL. PHOENIX14T is a DGS weather-related dataset with 8,257 signed videos and their corresponding gloss and German text. Its total vocabulary size is 1,115 glosses and 3,000 German text. The PHOENIX14T dataset is pre-divided into training, validation, and testing sets, whereas we split the MedASL dataset into 80% for 5-fold cross-validation and 20% for testing.

- Evaluation Metric

We measure the time efficiency by calculating the training and validation time (training time) in seconds. We also assess the accuracy using normalized BLEU score [54] with n-grams from 1 to 4, as shown in equations (5) and (6).

$$BLEU = BP \cdot e^{(\sum_{n=1}^{N} w_n \, log(p_n))} \quad (5)$$

where $p_n$ is the precision of n-grams, $w_n$ is the weight of each n-gram size, and $BP$ is the Brevity Penalty.

$$BP = \begin{cases} 1, & if\ c > r \\ e^{(1-\frac{r}{c})}, & if\ c \le r \end{cases} \quad (6)$$

where $c$ is the length of the candidate machine translation and $r$ is the reference corpus length.

## *5.2. Experiments*

We employ ADAT using TensorFlow on 2 NVIDIA RTX A6000 GPUs. To ensure efficient computation across datasets, we resize the input frames to 52×65, standardizing resolution without cropping operations.

We performed a structured multi-stage hyperparameter search for training ADAT to identify optimal values. In each stage, subsets of hyperparameters were tuned while others were fixed. The optimization objective was to maximize BLEU-4 on the validation set while focusing on key architectural and training parameters, as shown in Table 3. Throughout all experiments, we used sparse categorical cross-entropy with label smoothing of 0.1, Adam optimizer, learning rate schedule that reduced the rate by a factor of 0.5 down to $2\times10^{-6}$ with a patience of 9 epochs, and early stopping with a patience of 15 epochs based on the validation loss.

**Table 3:** Hyperparameter tuning for the transformer architectures understudy.

| Hyperparameter | Search Space | Optimal values for S2G2T | Optimal values for S2T |
|---|---|---|---|
| Number of encoders | 1, 2, 3, 4, 5, 6, 7, 8, 9, 10, 11, 12 | 12 | 1 |
| Number of decoders | 1, 2, 3, 4, 5, 6, 7, 8, 9, 10, 11, 12 | 12 | 1 |
| Hidden units | 256, 512, 1024 | 1024 | 512 |
| Number of heads | 4, 8, 16 | 16 | 8 |
| Feedforward size | 1024, 2048, 4096 | 1024 | 1024 |
| Dropout rate | 0, 0.1, 0.2, 0.4, 0.5, 0.6 | 0 | 0.1 |
| Learning rate | $10^{-3}$, $10^{-4}$, $10^{-5}$, $2\times10^{-5}$, $3\times10^{-5}$, $4\times10^{-5}$, $5\times10^{-5}$ | $5\times10^{-5}$ | $10^{-3}$ |
| Weight decay | 0, 0.1, $10^{-2}$, $10^{-3}$ | 0 | $10^{-3}$ |

NA: Not applicable; NR: Not reported.

## 6. Experimental Results Analysis

This section presents a comprehensive numerical evaluation of our proposed ADAT model on S2G2T and S2T translation tasks. We compare ADAT to several transformer-based baselines using BLEU scores for translation quality, training time for efficiency, and FLOPs for computational complexity. Table 4 summarizes the translation performance across both tasks and datasets.

**Table 4:** Comparison of models for sign-to-gloss-to-text and sign-to-text translations (scores range between 0 and 1, higher is better).

| Transformer Model | Dataset | Validation | | | | Test | | | |
|---|---|---|---|---|---|---|---|---|---|
| | | BLEU-1 | BLEU-2 | BLEU-3 | BLEU-4 | BLEU-1 | BLEU-2 | BLEU-3 | BLEU-4 |
| *Sign-to-gloss-to-text* | | | | | | | | | |
| Encoder-Decoder | PHOENIX14T | 0.370 | **0.214** | **0.141** | 0.096 | 0.373 | 0.219 | 0.145 | 0.099 |
| ADAT | | **0.371** | 0.210 | 0.139 | **0.097** | **0.374** | 0.219 | 0.145 | **0.1** |
| Encoder-Decoder | MedASL | 0.583 | 0.483 | 0.421 | 0.352 | **0.314** | **0.193** | **0.122** | **0.074** |
| ADAT | | **0.584** | **0.484** | **0.425** | **0.353** | 0.307 | 0.189 | 0.119 | 0.069 |
| *Sign-to-text* | | | | | | | | | |
| Encoder-Decoder | PHOENIX14T | 0.100 | 0.034 | 0.020 | 0.006 | 0.030 | 0.010 | 0.006 | 0.002 |
| Encoder-Only | | 0.133 | 0.052 | 0.034 | 0.025 | 0.130 | 0.060 | 0.042 | 0.032 |
| Decoder-Only | | 0.000 | 0.000 | 0.000 | 0.000 | 0.002 | 0.000 | 0.000 | 0.000 |
| ADAT | | **0.349** | **0.199** | **0.133** | **0.093** | **0.346** | **0.197** | **0.130** | **0.089** |
| Encoder-Decoder | MedASL | 0.610 | 0.513 | 0.447 | 0.383 | 0.311 | 0.193 | 0.123 | 0.074 |
| Encoder-Only | | 0.123 | 0.045 | 0.014 | 0.006 | 0.116 | 0.003 | 0.009 | 0.005 |
| Decoder-Only | | 0.061 | 0.008 | 0.003 | 0.002 | 0.037 | 0.005 | 0.003 | 0.001 |
| ADAT | | **0.636** | **0.551** | **0.490** | **0.430** | **0.317** | **0.202** | **0.123** | **0.076** |

### *6.1. Sign-to-Gloss-to-Text*

On PHOENIX14T, ADAT performs comparably to the encoder-decoder baseline, with marginal differences across all BLEU scores. In particular, ADAT achieves a slightly higher BLEU-4 score on the test set, indicating improved fluency in longer n-gram generation. However, on MedASL, a noticeable performance gap emerges between validation and test sets for both models, highlighting domain shift and generalization challenges. ADAT consistently improves validation scores, particularly in BLEU-3 and BLEU-4. However, it slightly underperforms on the test set, suggesting overfitting and reduced robustness on unseen data.

Fig. 2 presents the training times for both models. On PHOENIX14T, ADAT is 14.3% faster than the baseline, reducing training costs while maintaining translation quality. On MedASL, where overall training time is lower due to dataset size, ADAT still achieves a 3.2% speedup. These results confirm that ADAT offers competitive translation performance while being more computationally efficient.

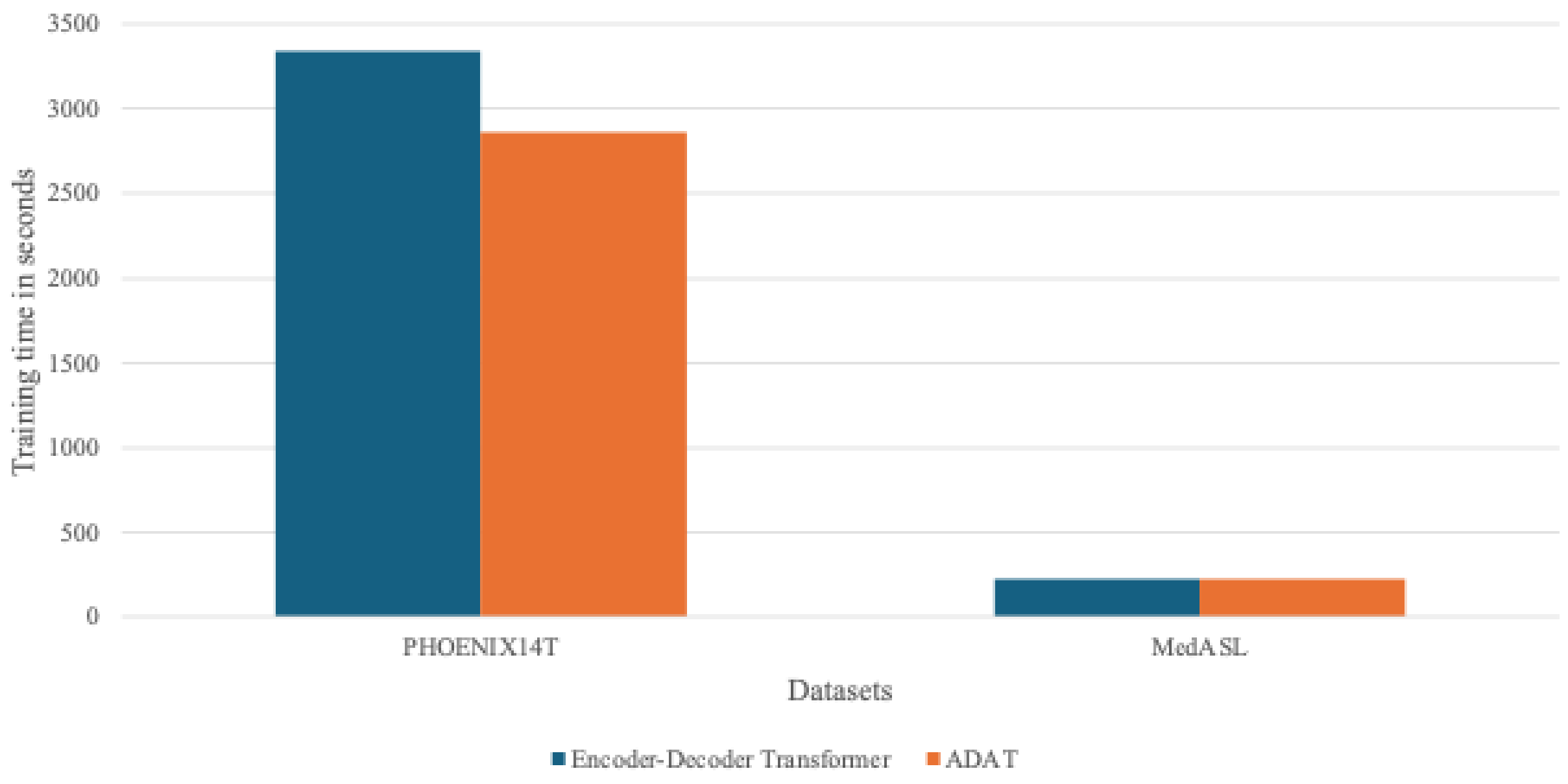

**Figure 2:** Sign-to-gloss-to-text training time.

In summary, in S2G2T translation, ADAT maintains comparable or better translation performance than the baseline while significantly reducing training time—especially on larger datasets—making it a more scalable and efficient choice.

*6.2. Sign-to-Text*

In direct S2T translation, ADAT outperforms encoder-decoder, encoder-only, and decoder-only baselines across both datasets. In particular, on PHOENIX14T, ADAT substantially improves BLEU-4 compared to the encoder-decoder baseline. On the other hand, encoder-only and decoder-only variants perform significantly worse, highlighting the importance of combining gloss grounding and language modeling in a unified architecture. On MedASL, ADAT surpasses the encoder-decoder baseline, while encoder-only and decoder-only variants achieve some degree of alignment, their performance remains significantly lower.

Fig. 3 shows that ADAT maintains competitive training efficiency. On PHOENIX14T, it is 2.7% faster than the encoder-decoder baseline. However, while the decoder-only trains faster, it fails to produce meaningful translations. On MedASL, ADAT maintains its advantage with a 7.2% reduction in training time, balancing efficiency and accuracy.

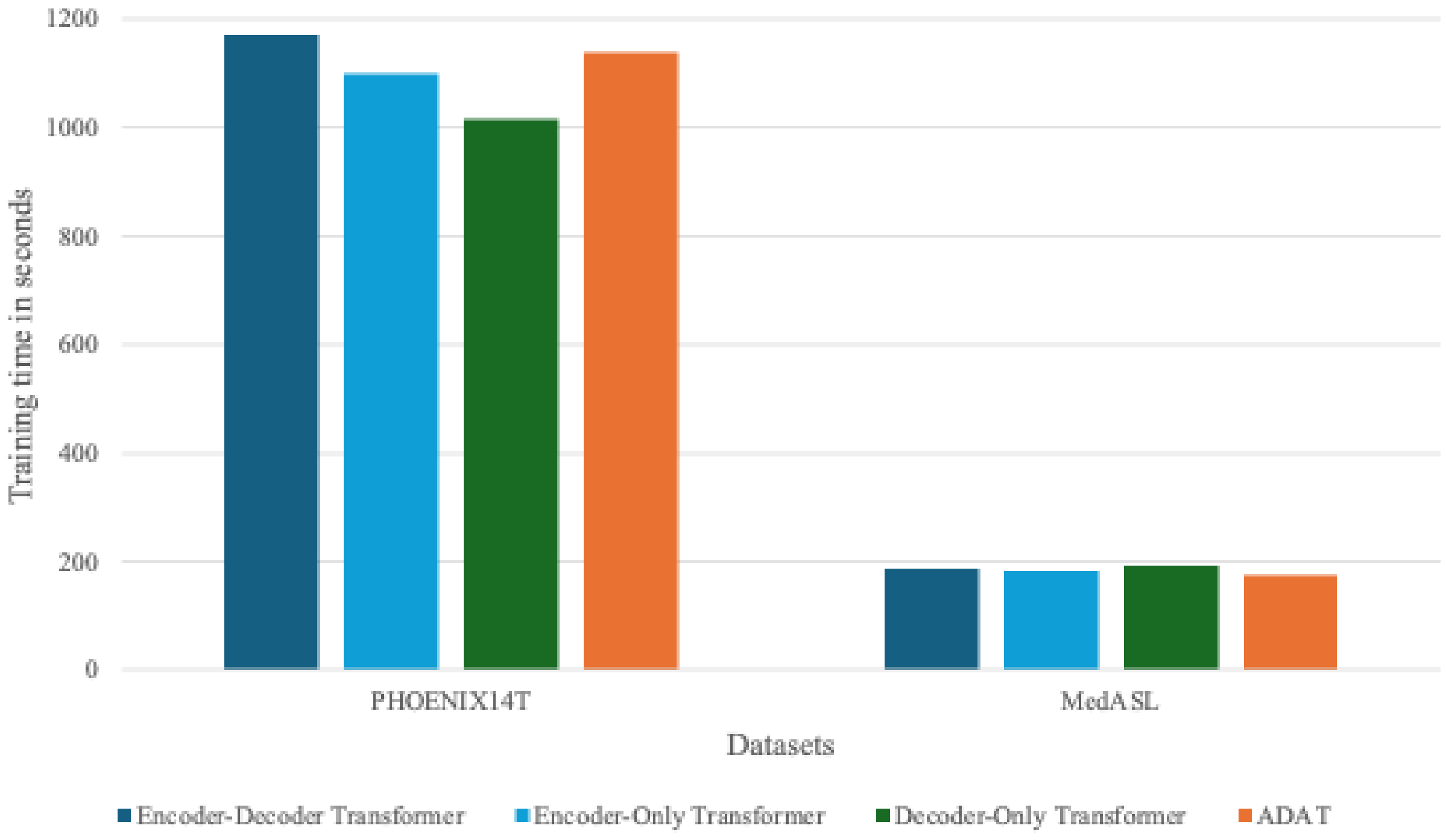

**Figure 3:** Sign-to-text training time.

In summary, ADAT achieves state-of-the-art translation quality on S2T tasks across both datasets while remaining competitively efficient to train, outperforming simplified transformer variants in both accuracy and generalizability.

To conclude, across both S2G2T and S2T tasks, ADAT consistently delivers strong translation performance while improving training efficiency relative to baseline models. These results demonstrate ADAT's ability to scale to larger datasets and maintain robustness across different domains, offering a practical and efficient solution for sign language translation.

### *6.3. Complexity Efficiency Analysis*

We further examine the computational efficiency of ADAT versus the encoder-decoder transformer by measuring the Floating-Point Operations Per Second (FLOPs), summarized in Table 5. Experiments were conducted on PHOENIX14T in a controlled experimental setup with 1 encoder, 1 decoder, 512 hidden units, 2048 feed-forward size, 0.1 dropout rate, and an initial learning rate of $5\times10^{-5}$, reduced by a factor of 0.5 down to a minimum of $2\times10^{-6}$. Our analysis reveals the impact of the encoder output length on the decoder complexity and, consequently, the overall computational performance in S2G2T and S2T.

In S2G2T, the encoder generates a gloss sequence consisting of 27 tokens, whereas in S2T, it produces a high-dimensional representation of the entire video sequence with 371 frames. This difference significantly impacts the decoder's computational complexity despite employing an identical decoder in both models. In particular, the shorter encoder output in S2G2T reduces the computational overhead during decoding, requiring only 1.85 Giga FLOPs (GFLOPs) in both models. However, the transformer encoder is the most computationally expensive, requiring 12.74 GFLOPs, whereas ADAT requires 6.72 GFLOPs. This optimization results in a 26.2% reduction in training time.

Conversely, S2T introduces a more significant computational burden due to the increased encoder output length. The decoder must process 371 frame sequences instead of 27 gloss representations, increasing the cross-attention workload and overall FLOPs. As a result, the decoding requires 5 GFLOPs, which is higher than S2G2T, leading to a significant computational overhead. While ADAT effectively reduces the encoder FLOPs from 12.74 to 2.08 GFLOPs, the overall training speedup is minor, with only a 6.8% reduction in training time. The increased decoder complexity limits the efficiency gains from encoder optimizations, making S2T more computationally expensive than S2G2T.

**Table 5:** Comparison of complexity efficiency.

| Translation | Sign-to-gloss-to-text | | Sign-to-text | |
|---|---|---|---|---|
| Model | Transformer | ADAT | Transformer | ADAT |
| Encoder Input Length | 371 | 371 | 371 | 371 |
| Decoder Input Length | Gloss: 27<br>Text: 52 | Gloss: 27<br>Text: 52 | Video: 371<br>Text: 52 | Video: 371<br>Text: 52 |
| Encoding FLOPs | 12.74 Giga | 6.72 Giga | 12.74 Giga | 2.28 Giga |
| Decoding FLOPs | 1.85 Giga | 1.85 Giga | 5 Giga | 5 Giga |
| Total FLOPs | 14.59 Giga | 8.57 Giga | 17.74 Giga | 7.28 Giga |
| Training Time | 2428.57 seconds | 1791.67 seconds | 1160.84 seconds | 1081.89 seconds |

In summary, S2G2T benefits from a shorter encoder output, reducing decoder complexity and significantly improving computational efficiency. ADAT's encoder optimization in S2G2T significantly decreases the FLOPs, resulting in a 26.2% reduction in training time. However, the more extended encoder output in S2T increases decoding complexity, diminishing the impact of ADAT's encoder. While ADAT achieves substantial FLOP reductions, the training time improvement is more pronounced in S2G2T due to the lower decoder computations.

## 7. Conclusion and Future Work

This study proposes a novel Adaptive Transformer (ADAT) to improve the Sign-to-Gloss-to-Text and Sign-to-Text translations by effectively capturing sign language's short- and long-range temporal dependencies. We validate ADAT using the RWTH-PHOENIX-Weather-2014 (PHOENIX14T) dataset and MedASL, a novel medical sign language dataset introduced in this study, as a benchmark for assessing translation accuracy in healthcare contexts. Our comparative evaluations show that ADAT outperforms baseline transformer models, including the encoder-decoder, encoder-only, and decoder-only, by significantly reducing training time while maintaining translation accuracy. These findings highlight ADAT's potential for creating efficient and scalable SLMT systems, bridging communication barriers, and fostering inclusivity for the Deaf community.

Future work will integrate multilingual components into ADAT architecture. In addition, we aim to develop a lightweight version of ADAT optimized for deployment on edge devices, facilitating real-world applications in resource-constrained environments where computational resources are constrained.

# Supplementary Information

## Supplementary 1: Adaptive Transformer (ADAT) Algorithm

**Algorithm 1** Adaptive Transformer (ADAT)

***Stage 1:*** *Feature Extraction*

**Input:** $x = \{x_f\}_{f=1}^{F}$
**Output:** $x_e$
**for** each $x_f$ **do**
 $CNN_{Features} = 2DCNN(x_f)$
 $Pooled_{Features} = MaxPooling(CNN_{Features})$
 $Flattened = Flatten(Pooled_{Features})$
 Append $Flattened$ to $x_e$

***Stage 2:*** *Encoder Processing*

**Input:** $x_e$
**Output:** $x_{e1}$, $x_{e2}$
Split $x_e$ equally into two halves along time dimension:
 $x_e = [x_{e1}^{F\times m_1 \times n_1}, x_{e2}^{F\times m_2 \times n_2}]$

***Stage 2.1:*** *Convolution Path*

**Input:** $x_{e1}$
**Output:** $Conv_{X_{e1}}$
$Conv_{X_{e1}} = Conv(x_{e1})$

***Stage 2.2:*** *LogSparse Self-Attention (LSSA) Processing*

**Input:** $x_{e2}$
**Output:** $LSSA_{x_{e2}}$
$\mathrm{Q} = x_{e2} \cdot W_q$ , $\mathrm{K} = x_{e2} \cdot W_k$ , $\mathrm{V} = x_{e2} \cdot W_v$
**for** each position p in $x_{e2}$ **do**
 $I_p^j = \{p - 2^{\lfloor log_2 p \rfloor}, p - 2^{\lfloor log_2 p \rfloor - 1}, \ldots, p - 2^0, p\}$
 $LSSA_p = Softmax((Q[I_p^j] \cdot K[I_p^j]^T)/\sqrt{d/2})$
$LSSA_{x_{e2}} = Concat(LSSA_p \mid p \in x_{e2})$

***Stage 2.3:*** *Adaptive Gating Mechanism*

**Input:** V, $Conv_{X_{e1}}$, $LSSA_{x_{e2}}$
**Output:** $y_e$(predicted gloss)
$GAP_{x_{e2}} = GlobalAveragePooling(V)$
$g = Softmax(w \cdot LSSA_{x_{e2}} + b)$
$Gating_{x_{e2}} = g \cdot LSSA_{x_{e2}} + (1 - g) \cdot GAP_{x_{e2}}$
$y_e = LayerNorm(Concat(Conv_{X_{e1}}, Gating_{x_{e2}}))$

***Stage 3:*** *Decoder Processing*

**Input:** $y_e$
**Output:** $y_d$
$x_{embed} = Embedding(y_e)$
$x_d = Positional\ Encoding(x_{embed})$
$y_d = TransformerDecoder(x_d)$

**Supplementary 2: Log-Sparse Self-Attention Complexity Analysis Theorem Proof**

***Theorem 1.*** Let $n$ be the sequence length. In canonical self-attention, the computational complexity is $O(L)$ [13]. The Log-Sparse Self-Attention (LSSA) [16] reduces this complexity to $O(L(logL)^2)$ by attending only to a logarithmic subset of past tokens.

***Proof.*** In the canonical multi-head self-attention, each position $p$ in the input sequence $L$ and embedding dimension $d$ attends to all positions through a dot product between each $L$ query $(L_q)$ and $L$ key $(L_k)$ pair. This leads to quadratic scaling $(L_q \times L_k)$ with time complexity of $O(L^2)$ per layer. For a transformer with $h$ attention heads, the total time complexity per layer scales as: $O(h \cdot L^2)$.

LSSA selects a logarithmic subset of past frames, where the selected indices follow a logarithmic spacing instead of attending to all previous tokens. Specifically, each cell $p$ attends to cells at indices defined by:

$$I_p^j = \{p - 2^{\lfloor log_2 p \rfloor}, p - 2^{\lfloor log_2 p \rfloor - 1}, \ldots, p - 2^0, p\}$$

where the spacing between attended cells increases exponentially, reducing the number of attended positions per cell to $O(LogL)$. This reduction in complexity significantly improves the model's efficiency, allowing it to handle higher sign video frame rates with lower computational overhead compared to standard transformer architectures.

**Supplementary 3: Proposed MedASL Dataset Processing**

*3.1 Prompt Engineering Design*

To create MedASL, we design and develop prompts using the following methodology:

- High-Level Prompt Structure

We design a high-level prompt to generate realistic medical conversations in the following format:

*"Generate a realistic medical interaction between a patient and a [doctor/nurse/pharmacist/technician] in a healthcare setting. The conversation should involve common symptoms, medical advice, and questions about treatments or prescriptions. Ensure the language is clear, professional, and appropriate for real-world scenarios."*

- Refinement Process

We refine the high-level prompt by dividing it into low-level prompts. This is to improve the generated sentences' coherence and relevance using low-level prompt variations such as:

*"Generate 10 medical-related statements that a nurse might say when checking a patient's vitals."*

*3.2 Data Pre-processing*

We record the sign videos using Intel RealSense at a resolution of 1280×800 and stored them in ".npy" format. To prepare the video data for training, we applied the following pre-processing steps:

- Video Frames Extraction: We sample the videos at 30 frames per second (fps) to maintain motion fidelity.
- Frames Concatenation: We concatenate video frames corresponding to each sentence into a continuous sequence to align with gloss annotations.
- Padding: We apply zero-padding to align frame lengths across all videos.

For the sign language gloss and spoken language text, we applied the following additional pre-processing steps:

- Building Vocabularies: We create unique vocabularies for gloss and text data, including a special token <UNK> to represent unknown words.
- Assigning Unique Indices: We assign a unique index to each word in the gloss and text data for better processing.
- Tokenizing: We tokenize the gloss and text sequences into individual units to enable efficient input representation.
- Padding: We apply zero-padding to the sequences, ensuring uniform lengths for batch processing.
- Adding Special Tokens: We add special tokens such as <sos> (start of sequence) and <eos> (end of sequence) to mark the sequence boundaries.
- Gloss Alignment: We map each gloss annotation to its corresponding spoken language sentence.
- Data serialization: We store the pre-processed gloss, text, and video data in a standardized ".pkl" format for efficient input loading during model training.

These pre-processing steps, summarized in Fig. S1, ensure the dataset is standardized, robust, and optimized for training and evaluation. Table S1 provides a detailed comparison of the raw and pre-processed data.

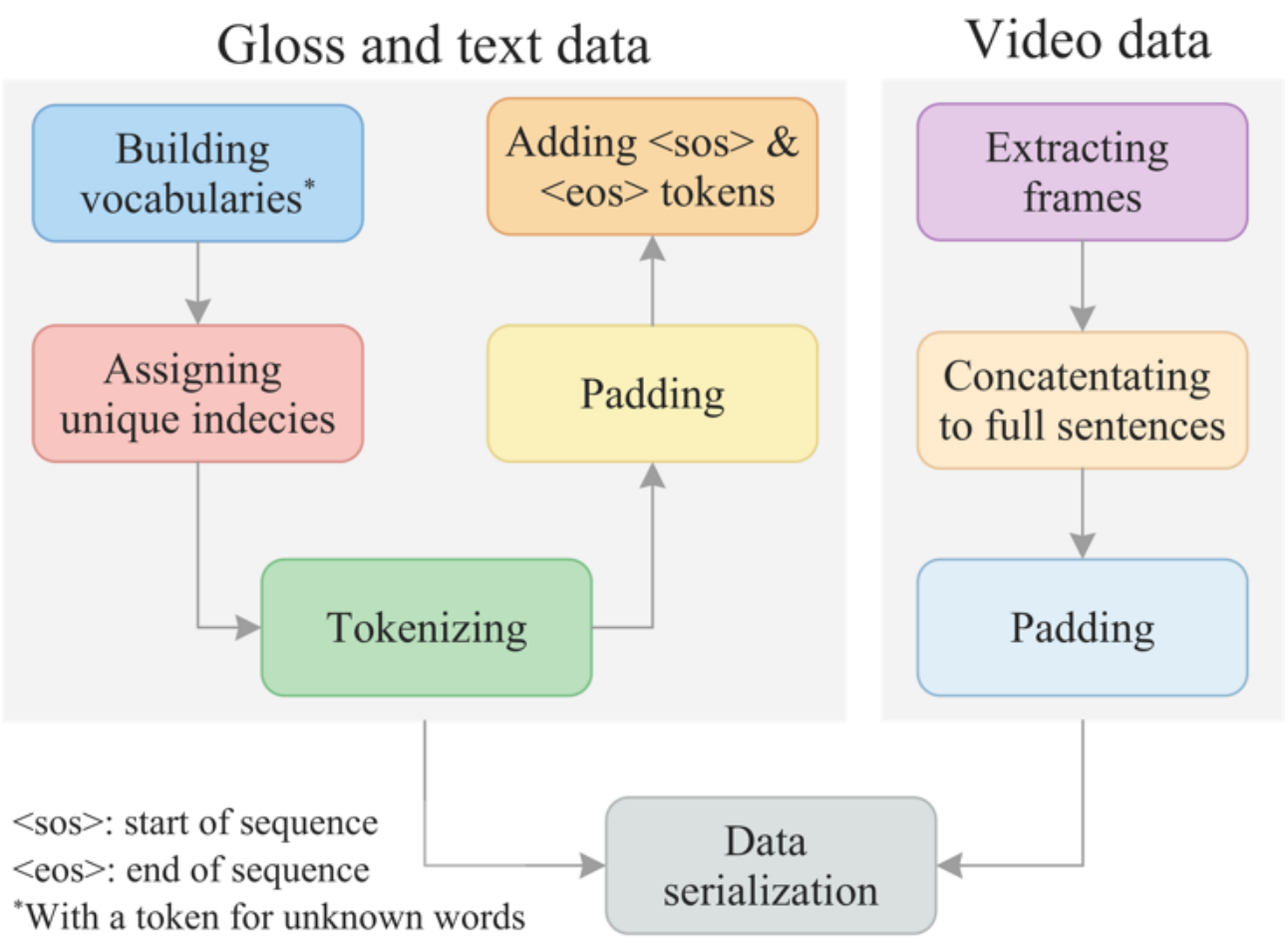

**Figure S1:** MedASL data pre-processing steps.

**Table S1:** Key statistics of the MedASL dataset.

| | Raw | Pre-processed |
|---|---|---|
| Gloss vocabulary size | 833 | 833 |
| Text vocabulary size | 912 | 912 |
| Total segments* | 29,793 | 59,000 |
| Total frames* | 893,790 | 1,770,000 |
| Total duration (in hours) | 8.3 | 16.4 |
| Max gloss length | 10 | 10 |
| Max text length | 16 | 16 |
| Max video length | 118 | 118 |
| #Signers | 1 | 1 |
| #Unique videos/sentences | 500 | 500 |